\documentclass{bmvc2k}
\usepackage{times}
\usepackage{epsfig}
\usepackage{graphicx}
\usepackage{amsmath}
\usepackage{amssymb}
\usepackage{times}
\usepackage{epsfig}
\usepackage{graphicx}
\usepackage{amsmath}
\usepackage{amssymb}
\usepackage{graphicx}
\usepackage{tabularx}
\usepackage{amsthm}
\newcolumntype{Y}{>{\centering\arraybackslash}X}
\usepackage{booktabs}
\usepackage{multirow}
\usepackage{float} 
\usepackage{multicol}
\usepackage{subfigure}
\usepackage[ruled,linesnumbered]{algorithm2e}
\usepackage{comment}

\usepackage{bm}

\title{DomainMix: Learning Generalizable Person \mbox{Re-Identification Without Human Annotations}}

\addauthor{Wenhao Wang*}{wangwenhao0716@gmail.com}{1}
\addauthor{Shengcai Liao$^\dagger$}{scliao@ieee.org}{2}
\addauthor{Fang Zhao}{fang.zhao@inceptioniai.org}{2}
\addauthor{Cuicui Kang}{cuicui.kang@mbzuai.ac.ae}{3}
\addauthor{Ling Shao}{ling.shao@ieee.org}{2,3}
\addinstitution{
 Baidu Research\\
 Beijing, China
}
\addinstitution{
 Inception Institute of Artificial Intelligence (IIAI)\\
Masdar City, Abu Dhabi, UAE
}
\addinstitution{
 Mohamed bin Zayed University of Artificial Intelligence (MBZUAI)\\
Masdar City, Abu Dhabi, UAE
}

\runninghead{Wang, Liao, etal}{DomainMix}
\def\eg{\emph{e.g}\bmvaOneDot}

\begin{document}

\maketitle
\vspace*{-5mm}
\begin{abstract}

Existing person re-identification models often have low generalizability, which is mostly due to limited availability of large-scale \textbf{\textit{labeled}} data in training. However, labeling large-scale training data is very expensive and time-consuming, while large-scale synthetic dataset shows promising value in learning generalizable person re-identification models. Therefore, in this paper a novel and practical person re-identification task is proposed, i.e. \ how to use \textbf{\textit{labeled}} synthetic dataset and \textbf{\textit{unlabeled}} real-world dataset to train a universal model.  In this way, human annotations are no longer required, and it is scalable to large and diverse real-world datasets. To address the task,  we introduce a framework with high generalizability, namely DomainMix. Specifically, the proposed method firstly clusters the unlabeled real-world images and selects the reliable clusters. During training, to address the large domain gap between two domains, a domain-invariant feature learning method is proposed, which introduces a new loss, i.e.  \ domain balance loss, to conduct an adversarial learning between domain-invariant feature learning and domain discrimination, and meanwhile learns a discriminative feature for person re-identification. This way, the domain gap between synthetic and real-world data is much reduced, and the learned feature is generalizable thanks to the large-scale and diverse training data. Experimental results show that the proposed annotation-free method is more or less comparable to the counterpart trained with full human annotations, which is quite promising. In addition, it achieves the current state of the art on several person re-identification datasets under direct cross-dataset evaluation.
\end{abstract}

\begin{figure}[t]  
\centering  
\includegraphics[width=0.95\textwidth]{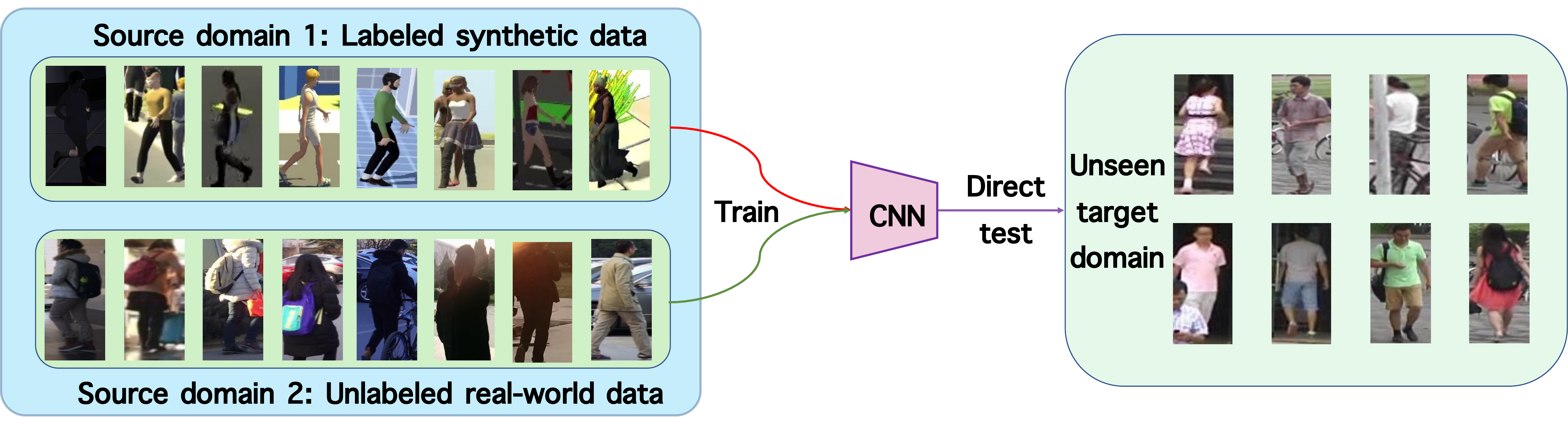}  
\caption{The illustration of the proposed task A+B$\rightarrow$C, i.e. \ how to use labeled synthetic data A and unlabeled real-world data B to train a model that can generalize well to an unseen target domain C.}  
\label{illu}  
\vspace*{-5mm}
\end{figure}
 \vspace*{-5mm}
\section{Introduction}
The goal of person re-identification (re-ID) is to match a given person across many gallery images captured at different times, locations, etc. With the development of deep learning, fully supervised person re-ID has been extensively investigated \cite{sun2018beyond, quan2019auto, luo2019strong, zheng2019joint, miao2019pose,liu2020deep,liu2020memory} and gained impressive progress. However, significant performance degradation can be observed when a trained model is tested on a previously unseen dataset. The generalizability of known algorithms is hindered by two main aspects. First, the generalizability of an algorithm is often ignored by its designer. There are only a few methods designed for domain generalization (DG). Second, the number of subjects in public datasets is limited, and their diversity is insufficient.  \par 

Labeling large-scale and diverse real-world datasets is expensive and time-consuming. For instance, labeling a dataset of the magnitude of MSMT17 \cite{wei2018person} requires three labelers to work for two months. To address this, RandPerson \cite{wang2020surpassing} inspires us to use large-scale synthetic data for effective person re-identification training, which gets rid of the need of human annotations. However, if using synthetic data alone, the generalizability of the learned model is still limited due to the domain gap between the synthetic and real-world data. Therefore, a solution is provided in \cite{wang2020surpassing} which learns from mixed synthetic data and labeled real-world data. However, though performance is improved, this solution still relies on heavy human annotations of the real-world data, and the domain gap still exists which is sub-optimal for generalization.

Therefore, the goal of this paper is to learn generalizable person re-identification completely without human annotations, so as to make use of a large amount of unlabeled real-world data. Specifically, we aim at how to combine a labeled synthetic dataset with unlabeled real-world data to learn a ready-to-use model with good generalizability. The proposed setting is illustrated in Fig. \ref{illu}, which is denoted as A (labeled) + B (unlabeled) $\rightarrow$ C (unseen target domain) with direct cross-dataset evaluation on C. The key to achieve domain generalization here is to make full use of the discriminative labels in the synthetic domain and the style and diversity of unlabeled real-world images simultaneously. A plausible method to tackle this problem would be Unsupervised Domain Adaptation (UDA) from A to B and trying to test it on C. However, the goal of UDA is different; it transfers the knowledge from the source domain A to the target domain B, and the testing is performed on the same target domain B. After the transfer, the model will learn domain-specific features from the less reliable real-world data without annotations and ignore the value of the large-scale high-quality labeled synthetic data. Therefore, directly applying UDA from A to B will have inferior generalizability on C. A task which may seem similar to the proposed one is the semi-supervised learning (SSL). However, for the SSL, both the labeled and unlabeled images are usually from the same domain while in the proposed setting, the images for training are from quite different domains. Besides, that is why we design special method to reduce the domain gap to improve generalizability.\par 

To address this problem, a solution called DomainMix is proposed, for discriminative, domain-invariant, and generalizable person re-identification feature learning. Specifically, to better utilize unlabeled real-world images, in each given epoch, they are clustered by DBSCAN algorithm. However, because unlike most UDA algorithms,  i.e. \ there is a pre-training on a labeled source dataset, the clustering results may be unreliable and noisy. Therefore, three criteria, i.e. \ independence, compactness and quantity, are used to select reliable clusters. After clustering in each epoch, the number of identities for training is various. Therefore, it is impossible to use the same classification layer all the time. To address the problem, an adaptive initialization method is utilized: The classification layer can be divided into two parts: one for the synthetic data and the other for the real-world data. The number of identities for the synthetic data part never changes, therefore, it is initialized as the result of the last epoch. However, for the real-world data part, the number of identities changes all the time. As a result, it is initialized as the average of the features of corresponding identity. This initialization method accelerates and guarantees the convergence of training. To deal with the huge domain gap between synthetic and real-world data, a domain-invariant feature learning method is designed. Through alternate training between backbone and discriminator, and with the help of the proposed domain balance loss and other person re-ID metrics, the network can learn discriminative, domain-invariant and generalizable features from two domains jointly. With this framework, the need of human annotations is completely eliminated, and the domain gap between the synthesized and real-world data is reduced, so that the generalizability is improved thanks to the large-scale and diverse training data.


The contributions are summarized as three-fold.
\begin{itemize}
  \item{The paper proposes a novel and practical person re-identification task, i.e. \ how to combine labeled synthetic dataset with unlabeled real-world dataset to train a model with high generalizability.}
  \item{A novel and unified DomainMix framework is proposed to learn discriminative, domain-invariant, and generalizable person re-identification feature from two domains jointly. For the first time, domain generalizable person re-identification can be learned without human annotations completely.}
  \item{Experimental results show the proposed annotation-free framework achieves comparable performance with its counterparts trained with full human annotations.}
\end{itemize}
 \vspace*{-5mm}
\section{Related Work}
\vspace*{-2mm}
\subsection{Unsupervised Domain Adaptation for Person Re-ID}
The goal of Unsupervised Domain Adaptation (UDA) for person re-ID is to learn a model on a labeled source domain and fine-tune it to an unlabeled target domain. The main UDA algorithms can be categorized into three classes. The first is image-level methods \cite{zhong2018generalizing, deng2018image, wei2018person, li2019cross}, which use a generative adversarial network (GAN) \cite{goodfellow2014generative} to translate the image style. The second class is feature-level methods \cite{li2018adaptation, chang2019disjoint, Lin2018MultitaskMF}, which aim to find domain-invariant features between different domains. The last category is cluster-based algorithms \cite{lin2019bottom, zhai2020ad, yang2020asymmetric, zeng2020hierarchical, ge2020mutual, ge2020selfpaced, fu2019self, zhao2020unsupervised,ding2020adaptive}, which generate pseudo labels to help fine-tune on the target domain. \par 
Although the UDA task and the proposed task both have the source and target domain, they are totally different. The goal of UDA is to use labeled source domain and unlabeled target domain to train a model which can perform well on the known target domain, while the proposed task aims to learn a model from labeled synthetic dataset and unlabeled real-world dataset to generalize well to an unseen domain.


\vspace*{-2mm}
\subsection{Domain Generalization for Person Re-ID}
Domain Generalization (DG) for person re-ID was first studied in \cite{yi2014deep}, aiming to generalize a trained model to unseen scenes. In recent years, with the increasing accuracy of fully supervised person re-ID and the limitations of UDA, DG has begun to attract attention again. For instance, DualNorm \cite{jiebmvc} uses instance normalization to filter out variations in style statistic in earlier layers to increase the generalizability. SNR \cite{jin2020style} filters out identity-irrelevant interference and keeps discriminative features by using an attention mechanism. QAConv \cite{LiaoQAConv} constructs query-adaptive convolution kernels to find local correspondences in feature maps, which is more generalizable than using features.  M$^3$T \cite{zhao2021learning} introduces meta-learning strategy and proposes a memory-based identification loss to enhance the generalization ability of the model. Other works, such as RandPerson \cite{wang2020surpassing}, focus on using synthetic data to enlarge the diversity and scale of person re-ID datasets. 
\vspace*{-2mm}
\subsection{Methods for Reducing Domain Gap}
Domain gap hinders one trained model performs well on an unseen dataset \cite{li2021DCC}. In the task of UDA for person re-ID, some methods, such as PTGAN \cite{wei2018person}, utilize GAN \cite{goodfellow2014generative} to transfer the image style of the source domain to the target domain. The methods reduce the domain gap from the image-level. Another category is feature-level and our method belongs to it. Some methods try to train a domain-invariant model by reducing the pairwise domain discrepancy with  Maximum Mean Discrepancy (MMD) \cite{tzeng2014deep}. However, this pipeline, which shares the same classes between domains, is not suitable for person re-ID task because the identities in two re-ID domains are different. 
\begin{figure}[t]  
\centering  
\includegraphics[width=1\textwidth]{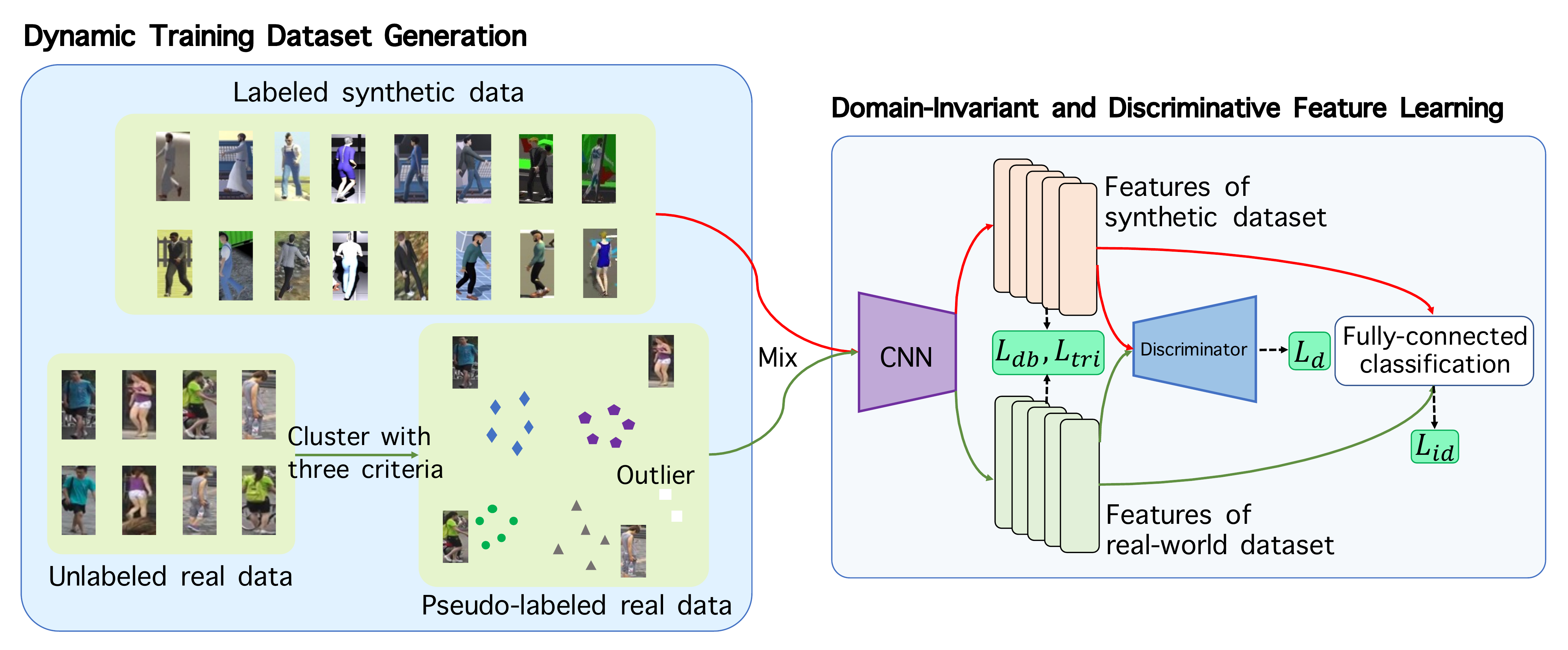}  
\\[3pt]
\caption{The design of the DomainMix framework.  During training, the backbone is trained to extract discriminative, domain-invariant, and generalizable features from two domains jointly with the help of the domain balance loss and other person re-identification metrics.}  
\label{Framework}  
\vspace*{-4mm}
\end{figure}
 \vspace*{-5mm}
\section{Proposed Task and Method}
\vspace*{-2mm}
\subsection{Problem Definition}
Two source domains $S_1$ and $S_2$, where $S_1$ is a synthetic dataset and $S_2$ is a real-world dataset, are given. For the synthetic dataset, the labels and images are both available. It is denoted as ${D_{{s_1}}} = \left\{ {\left( {x_i^{{s_1}},y_i^{{s_1}}} \right)\left| {_{i = 1}^{{N_{{s_1}}}}} \right.} \right\}$, where $x_i^{{s_1}}$ and $y_i^{{s_1}}$ are the $i$-th training sample and its corresponding person identity label, respectively, and $N_{{s_1}}$ is the number of images in the synthetic dataset. For the real-world dataset, only the images are available. The ${{N_{{s_2}}}}$ images in the real-world dataset are denoted as ${D_{{s_2}}} = \left\{ {x_i^{{s_2}}\left| {_{i = 1}^{{N_{{s_2}}}}} \right.} \right\}$. Besides, a target domain $T$, which is a real-world dataset different from ${D_{{s_2}}}$, is given. It is denoted as $D_t = \left\{ {x_i^t\left| {_{i = 1}^{{N_t}}} \right.} \right\}$, where $x_i^{{t}}$ denotes the $i$-th target-domain image and $N_t$ is the total number of target-domain images. This setting simulates the practical application scene, i.e. \ synthesizing labeled datasets is time-saving and cheap, while labeling a large-scale real-world dataset is time-consuming and expensive. Our goal is to design an algorithm that can be trained on the datasets ${D_{{s_1}}}$ and ${D_{{s_2}}}$, and then directly generalized to unseen ${D_{{t}}}$  without fine-tuning.

\vspace*{-2mm}
\subsection{DomainMix Framework}
To tackle the problem mentioned above, we propose the DomainMix framework. In this framework, reliable training dataset is generated dynamically according to three criteria, and before training, the classification layer is initialized adaptively to accelerate the convergence of identity classifier training. When training, together with discriminative metrics, a domain balance loss is proposed to help learning domain-invariant feature. As a result, the proposed DomainMix framework can generalize well to unseen target domains. The framework is shown in Fig. \ref{Framework}.
\vspace*{-2mm}
\subsubsection{Two Domains Mixing} 
\quad\textbf{Dynamic Training Dataset Generation}\par 
The training dataset for DomainMix framework is generated dynamically in each epoch. Given ${D_{{s_2}}}$, the reliable images are selected according to three criteria, i.e. \ independence, compactness, and quantity.  \par 
For independence and compactness, they are proposed in SpCL \cite{ge2020selfpaced} to judge whether a cluster is far away from others and whether the samples within the same cluster have small inter-samples distances. Together with the $eps$ parameter in DBSCAN \cite{ester1996density}, independence is realized by increasing $eps$ to figure out whether more examples are included into original cluster while compactness is realized by decreasing $eps$ to find whether a cluster can be split.  Please refer to DBSCAN  \cite{ester1996density} and SpCL \cite{ge2020selfpaced} to have a deeper understanding about the independence and compactness criteria. \par 
 For quantity, we argue that a reliable cluster should contain enough number of images which brings diversity. Further, if clusters with small number of images are selected, there will be too many classes to train an identity classifier well.   We denote the pseudo-labels set generated in one epoch as  $L_1 = \left\{\left.l_{i}\right|_{i=1} ^{M}\right\}$, where $l_i$ is the $i$-th pseudo label, and $M$ is the total number of pseudo labels. Given the bound $b$, labels with a total number of images below $b$ are discarded. Thus,  the refined pseudo-labels set is obtained as 
 \begin{equation}
  L_{2}=\left\{l_{i} \mid l_{i} \in L_{1}, S\left(l_{i}\right)\geq b \right\},
 \end{equation}
where $S\left(l_{i}\right)$ denotes the number of images belonging to the $i$-th pseudo label. Note that the quantity criteria is different from the $min\_samples$ parameter in DBSCAN \cite{ester1996density}: the quantity criteria handles the outliers and clusters with few images while $min\_samples$ parameter controls the core points selection in the process of clustering. Simply adjusting the $min\_samples$ parameter cannot bring similar improvement with quantity criteria. \par 
After images from ${D_{{s_2}}}$ are encoded to features, and features are clustered by a certain algorithm (\eg DBSCAN \cite{ester1996density}),  the generated clusters are selected by the three criteria. The ablation study part will show the proposed quantity criterion is the key to the outstanding performance while the criteria from \cite{ge2020selfpaced} only bring slight improvement. In conclusion, the images in reliable clusters are kept, pseudo labeled, and trained with ones from labeled synethetic dataset.  

\textbf{Adaptive Classifier Initialization}\par 
Because the training dataset is generated dynamically in each epoch, the number of classes is variant. It is impossible to use the same classification layer in each epoch and random initialization may bring non-convergence problems. As a result, an adaptive classifier initialization method is utilized to accelerate the training of identity classification.\par 
A classification layer can be formed as
\begin{equation}
y=W^Tx+ \bm b,
\end{equation}
where $x$ is a batch of features, $W$ is a matrix, and $\bm b$ is a bias which is set as $\bm 0$ for convenience. Given the number of classes $M$ in the generated training dataset and the dim of features $d$, the shape of matrix $W$ is $d\times M$. Because of the linear properties of matrix, $W$ can be written as $\left( W_{1},W_{2}\right)$ in blocks. $W_{1}$ is a matrix of shape $d\times N$ and $W_{2}$ is a matrix of shape $d\times (M-N)$, where $N$ is the number of classes in synthetic domain. \par
For $W_{1}$, because the classes of synthetic domain never changes during the different epochs, in a new epoch, it is initialized as the final result of the last epoch. For $W_{2}$, because clustering and selecting are performed in each epoch, $M$ changes all the time. Denote $W_{2}$ as $\left( w_{1},w_{2},...,w_{M-N}\right) $, and $w_i$ is initialized as
\begin{equation}
w_{i}=\frac{1}{K_i} \sum^{K_i}_{j=1} f_{j_i}\left( i=1,2,...,M-N\right),
\end{equation}
where $K_i$ is the number of images belonging to the $i$-th cluster under the current epoch, and $f_{j_i}$ is the feature of the $j$-th image in the cluster.\par
The advantage of this adaptive initialization method lies in two aspects. For the synthetic part, the initialization method enjoys the convenience and stability of fully-supervised learning. For the real-world part, after initialization, the probability of a given feature belongs to its class is much larger than other classes, therefore training the classifier is much easier.
\vspace*{-2mm}

\subsubsection{Domain-Invariant and Discriminative Feature Learning}
Given the generated training dataset and a well-initialized network, this section focuses on how to learn discriminative, domain-invariant, and generalizable features from two domains. It is realized by training a discriminator and backbone alternately. The discriminator is used to classify a given feature into its domain. Specifically, features of the images from the synthetic and real-world domains are extracted by the backbone. Then a discriminator is trained to judge which domain the extracted feature comes from. When training the discriminator, the cross-entropy loss ${\cal L}_{ce}$ is adopted. Thus the domain classification loss is defined as 
\begin{equation}\label{wf3}
{\cal L}_{d}^{s}(\theta)=\frac{1}{N_{s}} \sum_{i=1}^{N_{s}} {\cal L}_{c e}\left(C_{d}\left(F\left(x_{i}^{s} \mid \theta\right)\right), d_{i}^{s}\right),
\end{equation}
where $F\left( { \cdot \left| \theta  \right.} \right)$ is a feature encoder function, $N_s$ is the sum of the number of images in the current generated dataset, $C_d$ denotes the discriminator and $d_i^s$ is the domain label of the $i$-th image, i.e. \ if the image belongs to the synthetic domain, $d_i^s = 0$, otherwise if it belongs to the real-world domain, $d_i^s = 1$. \par 
To encourage the backbone to extract domain-invariant features, it is trained to confuse the domain discriminator. Therefore, a domain balance loss is proposed, which is defined as
\begin{equation}
\mathcal{L}_{d b}=\frac{1}{N_{s}} \sum_{i=1}^{N_{s}}\left(\sum_{j=1}^{n}\left(x_{j}^{i} \log \left(x_{j}^{i}\right)+a\right)\right),
\end{equation}
where ${x_j^i}$ is the $j$-th coordinate of $C_{d}\left(F\left(x_{i}^{s} \mid \theta\right)\right)$, and $a$ is a constant to prevent a negative loss. In this loss, considering the function
\begin{equation}
f(x)=x \log (x)+a, x \in(0,1),
\end{equation}
the second derivative of $f$ is
\begin{equation}
f''\left( x \right) = \frac{1}{x} > 0.
\end{equation}

Therefore, it is a convex function. Given $\sum\nolimits_{j = 1}^n {x_j^i = } 1$, the minimum value of the function can be achieved when $x_{j}^{i}=1 / n(j=1,2, \ldots, n)$, according to Jensen's inequality. \par 
In conclusion, when ${\cal L}_{db}$ is minimized, the distance between $x_{j}^{i}$ and $1/n$ is shortened. Thus, the probability of a given feature belonging to two domains tends to be the same, i.e. \ the backbone can extract domain-invariant features by confusing the discriminator.\par 
Beyond learning domain-invariant features, the network is also trained by discriminative metrics in re-ID, therefore an identity classification loss $L_{id}^s\left( \theta  \right)$ and a triplet loss $L_{tri}^s\left( \theta  \right)$ \cite{hermans2017defense} are adopted. They are defined as

\begin{equation}
{\cal L}_{id}^s(\theta ) = \frac{1}{{{N_s}}}\sum\limits_{i = 1}^{{N_s}} {{{\cal L}_{ce}}} \left( {{C_s}\left( {F\left( {x_i^s\mid \theta } \right)} \right),y_i^s} \right),
\end{equation}
and
\begin{equation}
\small
\begin{array}{l}
{\cal L}_{t r i}^{s}(\theta)=\frac{1}{N_{s}} \sum_{i=1}^{N_{s}} \max \left(0, m+\left\|F\left(x_{i}^{s} \mid \theta\right)-F\left(x_{i, p}^{s} \mid \theta\right)\right\|\right. \\
\left.-\left\|F\left(x_{i}^{s} \mid \theta\right)-F\left(x_{i, n}^{s} \mid \theta\right)\right\|\right),
\end{array}
\end{equation}
where $C_s$ is an identity classifier, $\|\cdot\|$ denotes the $L^{2}$-norm distance, $m$ is the triplet distance margin, ${\cal L}_{ce}(\cdot, \cdot)$ represents the cross-entropy loss, $y_i^s$ is the corresponding label or generated label, and the subscripts $_{i,p}$ and $_{i,n}$ indicate the hardest positive and the hardest negative index for the sample $x_i^s$ in a mini-batch.\par 
Therefore, the final loss is calculated as 

\begin{equation}\label{wf2}
{\cal L}^{s}(\theta)=\lambda^{m} {\cal L}_{db}(\theta)+\lambda^{s} {\cal L}_{i d}^{s}(\theta)+{\cal L}_{t r i}^{s}(\theta),
\end{equation}
where $\lambda^{m}$ and $\lambda^{s}$ are the balance parameters. Through alternate training with ${\cal L}_{d}^{s}(\theta)$ and ${\cal L}^{s}(\theta)$, the discriminator can classify a given feature into its domain, and the backbone can extract domain-invariant and discriminative features.
To summarize the proposed algorithm, the pseudo codes are given in supplemental material.

\section{Experiments}
\subsection{Datasets and Evaluation Metrics}
To evaluate the generalizability of the proposed DomainMix framework, extensive experiments are conducted on four widely used public person re-ID datasets. Among them, RandPerson  (RP) \cite{wang2020surpassing} is selected as the synthetic dataset. Its subset contains $8,000$ persons in $132,145$ images. Nineteen cameras were used to capture them under eleven scenes.  All images in the subset are used as training data, i.e. \ no gallery or query is available. The real-world datasets used are Market-1501 \cite{zheng2015scalable}, CUHK03-NP \cite{zhong2017re, li2014deepreid}, and MSMT17 \cite{wei2018person}.  Note that DukeMTMC \cite{zheng2017unlabeled} dataset is not used due to the invasion of privacy.  The details of real-world dataset are illustrated in the supplemental material.
Evaluation metrics are mean average precision (mAP) and cumulative matching characteristic (CMC) at rank-$1$.
\subsection{Implementation Details}

DomainMix is trained on four Tesla-V100 GPUs. The ImageNet-pre-trained \cite{deng2009imagenet} ResNet-50 \cite{he2016deep} and IBN-ResNet-50 \cite{pan2018two} are adopted as the backbone. Adam optimizer is used to optimize the networks with a weight decay of $5 \times 10^{-4}$.  For more details, please refer to supplemental results.
\subsection{Ablation Study}
Comprehensive ablation studies are performed to prove the effectiveness of each component in the DomainMix framework. Two different DG tasks are selected: labeled RandPerson \cite{wang2020surpassing} with unlabeled MSMT17 \cite{wei2018person} to Market-1501 \cite{zheng2015scalable} and labeled RandPerson \cite{wang2020surpassing} with unlabeled CUHK03-NP \cite{zhong2017re, li2014deepreid} to Market-1501 \cite{zheng2015scalable}. The experimental results on ResNet-50 \cite{he2016deep} are reported below, and the results on IBN-ResNet-50 \cite{pan2018two} are shown in supplemental results.\par

\begin{table}
\caption{Ablation studies for each component in the DomainMix framework on the two tasks. `+I/C/Q' denotes the independence/compactness/quantity criteria is used. With or without ACI/DB denotes whether using adaptive classifier initialization/domain balance loss or not.  `Labeled' or `unlabeled' denotes whether real-world source training data is labeled or not.}
\vspace*{2mm}
  \footnotesize
  \begin{tabularx}{\hsize}{p{3.1cm}|YY|p{3.1cm}|YY}
    \hline
RP$+$MSMT $\to$ Market & mAP & rank-$1$ & RP$+$CUHK $\to$ Market & mAP & rank-$1$ \\
    \hline\hline
    DBSCAN & $37.5$ & $64.6$ &DBSCAN & $34.5$ & $61.3$  \\ 
    DBSCAN + I + C& $37.0$ & $64.2$ &DBSCAN + I + C& $35.5$ & $62.8$ \\ 
    DBSCAN + Q& $42.4$ & $69.4$ &DBSCAN + Q& $39.5$ & $66.2$ \\
    DBSCAN + I + C + Q& $43.5$ & $70.2$ &DBSCAN + I + C + Q& $39.8$ & $67.5$ \\ 
    \hline
    Without ACI& $29.5$ & $56.9$ &Without ACI& $33.8$ & $60.3$ \\
    With ACI& $43.5$ & $70.2$ &With ACI& $39.8$ & $67.5$ \\ 
    \hline
    Without DB& $40.1$ & $68.1$ &Without DB& $37.3$ & $66.0$ \\ 
    With DB& $43.5$ & $ 70.2$ &With DB& $39.8$ & $67.5$ \\ 
    \hline
    Only RandPerson & $36.5$ & $63.6$ & Only RandPerson & $36.5$ & $63.6$\\   
   Only MSMT (labeled) & $32.7$ & $62.0$ &Only CUHK (labeled) & $25.1$ & $50.3$ \\
    DomainMix (labeled) & $45.2$ & $70.5$  &DomainMix (labeled) & $42.7$ & $69.7$  \\
    DomainMix (unlabeled) & $43.5$ & $70.2$ &DomainMix (unlabeled) & $39.8$ & $67.5$ \\ 
    \hline
    
  \end{tabularx}
  \\

  \label{ABLLL}
\end{table}

\textbf{Effectiveness of Dynamic Training Dataset Generation.} To investigate the necessity of generating training dataset dynamically and the importance of each component, we compare the domain generalizability of a model trained on two different real-world datasets,  i.e. \ MSMT17 \cite{wei2018person} and CUHK03-NP \cite{zhong2017re, li2014deepreid}. The baseline model performances are shown in Table \ref{ABLLL} as ``DBSCAN". If the independence and compactness criteria are used, the performances are denoted as ``DBSCAN + I + C", while if the quantity criterion is used, they are denoted as ``DBSCAN + Q". ``DBSCAN + I + C + Q" denotes all the three criteria are adopted. The quantity criterion brings $4.9\%$ in mAP improvement for the ``RP$+$MSMT $\to$ Market'' task. For the ``RP$+$CUHK $\to$ Market'' task, the mAP increases by $5.0\%$. However, if the independence and compactness criterion are used alone, no stable performance improvement can be observed. It is because, although the two criteria remove the unreliable clusters, there are still many classes including few images to participate in the training, which disturbs the training of the identity classifier and leads to the failure to improve the performance stably. Together with the proposed quantity criterion, the above problem is solved, and the two criteria in \cite{ge2020selfpaced} can further improve the performance.\par 
\textbf{Effectiveness of Adaptive Classifier Initialization.} To prove the effectiveness of the adaptive classifier initialization method, the experimental results without and with this method are shown in Table \ref{ABLLL} and denote as ``Without ACI" and ``With ACI", respectively. The initialization method brings significant improvement of $14.0\%$ and $6.0\%$ in mAP on the ``RP$+$MSMT $\to$ Market'' and ``RP$+$CUHK $\to$ Market'' tasks. The significant improvement comes from the guarantee and acceleration of the convergence.\par 

\textbf{Influence of Domain Balance Loss.} To verify the necessity of using the domain balance loss to learn domain-invariant features, results obtained with and without this loss are compared and shown in Table \ref{ABLLL} as ``Without DB" and ``With DB", respectively. All experiments with the use of domain balance loss show distinct improvement on both the ``RP$+$MSMT $\to$ Market'' and ``RP$+$CUHK $\to$ Market'' tasks. Specifically, the mAP increases by $3.4\%$ when the real-world source domain is MSMT17 \cite{wei2018person}. As for the `RP$+$ CUHK $\to$ Market' task, similar mAP improvement of $2.5\%$ can be observed. The improvement brought by domain balance loss on CUHK is displayed in the supplemental material. \par
We further discuss the \textbf{importance of introducing unlabeled real-world dataset}, \textbf{whether human annotations are essential for generalizable person re-ID}, and the \textbf{comparison with UDA algorithms} in the supplemental material. 
\subsection{Comparison with the State-of-the-arts}
The proposed DomainMix framework is compared with state of the art methods on three DG tasks,  i.e. \ directly testing on Market1501 \cite{zheng2015scalable}, CUHK03-NP \cite{zhong2017re, li2014deepreid}, and MSMT17  \cite{wei2018person}. The experimental results are shown in Table \ref{SOTA}. Note that a fair comparison in anyway is not very feasible, because we only used unlabeled real-world data, although with additional synthesized data, while others used labeled one. So existing results in Table  \ref{SOTA} are only provided as a reference to see what we can achieve with a fully annotation-free setting. Secondly, the proposed method is orthogonal to network architecture designs such as IBN-Net \cite{pan2018two} and OSNet-IBN \cite{zhou2019omni}. Thus they can also be applied into the framework.  The related experimental results are shown in the supplemental material. For QAConv \cite{LiaoQAConv}, though its performance is relatively high, because it needs to store feature maps of images rather than features to match, more memory is needed. SNR \cite{jin2020style} uses attention mechanism to  solve the drawback of instance normalization and improve the performance of IBN-Net \cite{pan2018two}, and the DomainMix may achieve further performance improvement with the help of this plug-and-play module. \par 
From the comparison in Table \ref{SOTA}, the DomainMix framework improves up to $7.0\%$ mAP. The improvement in performance is attributed to two aspects. First, directly combining the training of the synthetic dataset and unlabeled real-world dataset increases the source domain's diversity and scale. Second, the domain balance loss further forces the network to learn domain-invariant features and minimizes the domain gap between the synthetic dataset and real-world dataset in the source domain.
\begin{table}
  \footnotesize
\caption{Comparison with state-of-the-arts on Market1501 \cite{zheng2015scalable}, CUHK03-NP \cite{zhong2017re, li2014deepreid}, and  MSMT17 \cite{zhong2017re, li2014deepreid}. `$^\S$' denotes the results are from github of the original paper, `$^\ast$' denotes our implementation, and `$^\dagger$' indicates that the results are reproduced based on the authors' codes. `L' or `U' denotes whether the real-world source training data is labeled or not. }
\vspace*{2mm}
  \footnotesize
  \begin{tabularx}{\hsize}{|p{2.82cm}|p{2.1cm}|YY|}
    \hline
    \multicolumn{1}{|c|}{\multirow{2}{*}{Method}} &
    \multicolumn{1}{c|}{\multirow{2}{*}{Source data}} &
    \multicolumn{2}{c|}{Market1501}  \\
    \cline{3-4}
      & & \footnotesize{mAP} &\footnotesize{rank-$1$} \\
    \hline\hline
    MGN \cite{wang2018learning,yuan2020calibrated}&MSMT (L)&$25.1 $ & $48.7 $   \\
    ADIN \cite{yuan2020calibrated}&MSMT (L)&$22.5 $ & $50.1 $   \\
    ADIN-Dual \cite{yuan2020calibrated} &MSMT (L)&$30.3 $ & $59.1 $ \\   
    SNR \cite{jin2020style} &MSMT (L)&$41.4$ & $70.1$ \\
    QAConv$^\dagger$ \cite{LiaoQAConv} &MSMT (L)& $35.8$ & $66.9$   \\
    \hline
    MGN$^\dagger$ \cite{wang2018learning}&RandPerson&$17.7$ & $37.4$   \\
    OSNet-IBN$^\dagger$ \cite{zhou2019omni} &RandPerson&$39.0$ & $67.0$ \\
    QAConv$^\S$ \cite{LiaoQAConv} &RandPerson& $34.8$ & $65.6$   \\
    Baseline$^\ast$ &RandPerson& $36.5$ & $63.6$   \\
    \hline
    DomainMix &RP$+$MSMT (U) &$43.5$ & $70.2$ \\
    \footnotesize{DomainMix-OSNet-IBN} &RP$+$MSMT (U)&$\textbf{44.6}$ & $\textbf{72.9}$  \\
    \hline
  \end{tabularx}

  \footnotesize
 \begin{tabularx}{\hsize}{|p{2.82cm}|p{2.1cm}|YY|}
    \hline
    \multicolumn{1}{|c|}{\multirow{2}{*}{Method}} &
    \multicolumn{1}{c|}{\multirow{2}{*}{Source data}} &
    \multicolumn{2}{c|}{CUHK03-NP}  \\
    \cline{3-4}
     & & \footnotesize{mAP} &\footnotesize{rank-$1$} \\
    \hline\hline
    MGN \cite{wang2018learning,qian2019leader} &Market (L)& $7.4$ & $8.5$  \\
    MuDeep \cite{qian2019leader} &Market (L)& $9.1$ & $10.3$  \\
    QAConv$^\dagger$ \cite{LiaoQAConv}&MSMT (L) & $15.2$ & $16.8$ \\
	\hline    
	MGN$^\dagger$ \cite{wang2018learning}&RandPerson&$7.7$ & $7.4$   \\
	OSNet-IBN$^\dagger$ \cite{zhou2019omni} &RandPerson&$12.9$ & $13.6$ \\
	QAConv$^\S$ \cite{LiaoQAConv} &RandPerson& $11.0$ & $14.3$   \\
    Baseline$^\ast$ &RandPerson& $13.0$ & $14.6$ \\
    \hline
    DomainMix&RP$+$MSMT (U) & $16.7 $ & $\textbf{18.0}$ \\
    \footnotesize{DomainMix-OSNet-IBN} &RP$+$MSMT (U)&$\textbf{16.9}$ & $17.5$  \\
    \hline
  \end{tabularx}
  \\

  \footnotesize
  \begin{tabularx}{\hsize}{|p{2.82cm}|p{2.1cm}|YY|}
    \hline
    \multicolumn{1}{|c|}{\multirow{2}{*}{Method}} &
    \multicolumn{1}{c|}{\multirow{2}{*}{Source data}} &
    \multicolumn{2}{c|}{MSMT17}  \\
    \cline{3-4}
      & & \footnotesize{mAP} &\footnotesize{rank-$1$} \\
    \hline\hline
    QAConv$^\dagger$ \cite{LiaoQAConv}&Market (L) & $8.3$ & $26.4$   \\
    \hline
    MGN$^\dagger$ \cite{wang2018learning}&RandPerson&$3.0$ & $10.1$   \\
    OSNet-IBN$^\dagger$ \cite{zhou2019omni} &RandPerson&$12.4$ & $34.3$ \\
    QAConv$^\S$ \cite{LiaoQAConv} &RandPerson& $10.7$ & $34.3$   \\
    Baseline$^\ast$ &RandPerson& $7.9$ & $23.0$  \\
    \hline
    DomainMix&RP$+$Market (U) & $9.3$ & $25.3$  \\
     \footnotesize{DomainMix-OSNet-IBN}&RP$+$Market (U) & $\textbf{13.6}$ & $\textbf{36.2}$  \\
    \hline
    
  \end{tabularx}
  \label{SOTA}
\end{table}

\vspace*{-5mm}
\section{Conclusion}
 
In this paper, a more practical and generalizable person re-ID task is proposed, i.e. \ how to combine a labeled synthetic dataset with unlabeled real-world data to train a more generalizable model. To deal with it, the DomainMix framework is introduced, with which the requirement of human annotations is completely removed, and the gap between synthesized and real-world data is reduced. Extensive experiments show that the proposed annotation-free method is superior for generalizable person re-ID.

\section*{Acknowledgements}
 The authors would like to thank  Anna Hennig who helped proofreading the paper.

\bibliography{egbib}
\end{document}